\documentclass[11pt]{article}

\usepackage[final]{acl}

\usepackage{times}
\usepackage{latexsym}

\usepackage[T1]{fontenc}

\usepackage[utf8]{inputenc}

\usepackage{microtype}

\usepackage{inconsolata}

\usepackage{graphicx}

\usepackage{amsmath} 
\usepackage[ruled,vlined]{algorithm2e}
\DontPrintSemicolon
\usepackage{multirow}
\usepackage{graphicx}
\usepackage{colortbl}
\usepackage{bm}
\usepackage{booktabs}
\usepackage{amsfonts}
\usepackage[table]{xcolor}

\title{Beyond Semantics: Modeling Factual and Affective Perceptual Experiences from Vision-Language Data}

\author{Youssef Mohamed\thanks{Corresponding Authors} \\
  KAUST \\
  \small{youssef.mohamed@kaust.edu.sa} \\\And
  Kenneth Ward Church \\
  VecML \\
  \small{kenneth.ward.church@gmail.com} \\\And
  Mohamed Elhoseiny$^*$ \\
  KAUST \\
  \small{mohamed.elhoseiny@kaust.edu.sa} 
 \\
  }

\begin{document}


\makeatletter
    \let\@oldmaketitle\@maketitle
    \renewcommand{\@maketitle}{\@oldmaketitle
        \myfigure\bigskip}
    \makeatother
    \newcommand\myfigure{%
    \vspace{-1cm}
      \includegraphics[width=\linewidth]{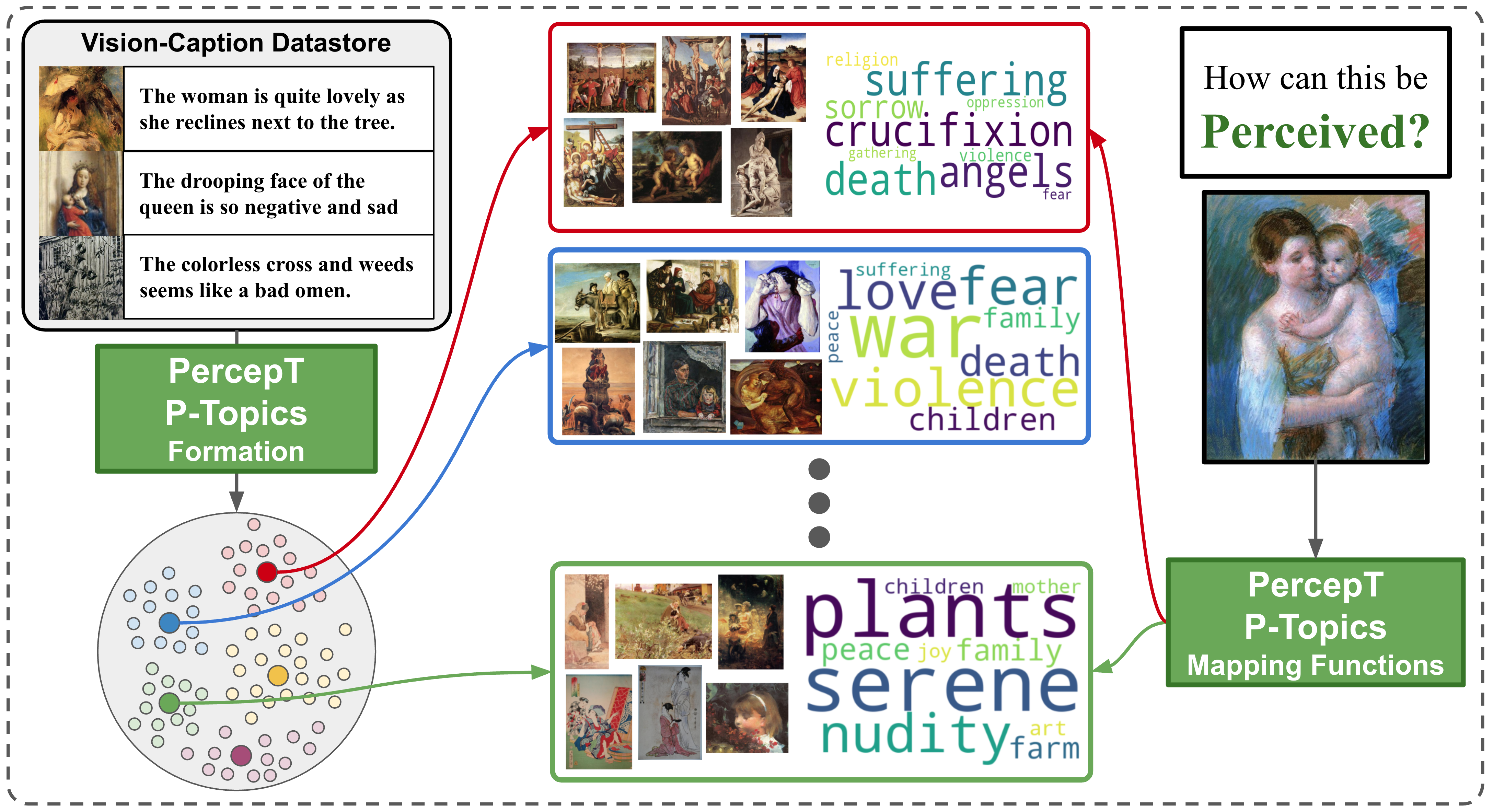}
        \refstepcounter{figure}\normalfont{Figure~\thefigure: 
        \textbf{Overview of PercepT.} PercepT clusters image-caption pairs into P-Topics capturing both factual and affective characteristics (left): religion/crucifixion (red), war/negative emotions (blue), and serenity/family (green). It then learns \emph{P-Topic mapping functions} that associate a new image (right) with its relevant P-Topics.
          }
      \label{fig:teaser}
    }

\maketitle

\begin{abstract}
We present P-Topics (Perception Topics) modeling, a novel problem for understanding how images are perceived affectively and across cultures. The goal is to (1) discover and model the different perception experiences in a dataset of images and captions, where each experience is defined by an objective factual and a subjective affective aspect, and (2) associate images to their relevant perception experiences.
We introduce \textbf{PercepT} (\textbf{Percep}tion topic \textbf{T}ransformer), a two-stage architecture that tackles P-Topics modeling. In the formation stage, percepT discovers \emph{P-Topics} as visual-textual clusters using an unsupervised training objective, and dynamically selects the number of clusters to match the perceptual richness of the dataset. In the mapping stage, it learns \emph{P-Topic mapping functions} via attention pooling to associate images to their respective clusters.
On ArtELingo, PercepT achieves a silhouette score of \textbf{0.97} compared to \textbf{0.37} from the closest baseline reflecting better perceptual clusters. PercepT also achieves an AUC score of \textbf{0.94} compared to \textbf{0.77} showing better mapping to perceptual clusters. Human evaluation confirms that PercepT captures semantically meaningful perception experiences and significantly outperforms existing methods. Our implementation will be made public.
\end{abstract}


\section{Introduction}
\label{sec:intro}

People from different cultural backgrounds respond differently to visual stimuli. For example, a painting of a mother and child (Figure~\ref{fig:teaser}) may evoke a serene family portrait for some, religious themes of Mary and Jesus for others, or fear and violence for those who connect the stern expressions to personal or historical contexts. These differing perceptions illustrate intricate interactions among images, language, and culture that this work aims to model.

We formalize this challenge by introducing \textbf{P-Topic modeling}: (1) discover the distinct \emph{perception experiences} in a dataset of image-caption pairs, decomposed into an \textbf{objective factual} and a \textbf{subjective affective} aspect, and (2) learn \textbf{P-Topic mapping functions} that associate unseen images with the discovered experiences. As shown in Figure~\ref{fig:teaser}, the same image activates distinct P-Topics — e.g., a religious P-Topic (red) connecting the mother and child to Mary and Jesus, and a serene P-Topic (green) emphasizing light colors and family warmth.

Our work sits at the intersection of text-based topic modeling \cite{blei2003latent, meng2022topic, Grootendorst2022BERTopicNT, Angelov2020Top2VecDR, sia-etal-2020-tired}, vision-language understanding \cite{elhoseiny2013write,elhoseiny2016write,Yang2022LanguageIA, Kazmierczak2023CLIPQDAAE, radford2021learning}, and cross-cultural VL research \cite{mohamed2024no, mohamed2022artelingo, mohamed2022okay}. While traditional topic models focus on text and vision-language models target general understanding, P-Topics specifically models the diverse perceptual perspectives that different people bring to the same visual content.

We propose \textbf{PercepT} (\textbf{Percep}tion topic \textbf{T}ransformer), a two-stage architecture combining DEC-based unsupervised clustering over fused factual and affective embeddings to discover and model distinct perceptual experiences as clusters. In stage 2, PercepT utilize attention-pooling to learn P-Topic mapping functions. Our method is detailed in \S\ref{sec:method}.

We utilize a CLIP-based encoder to represent the objective factual information and an emotion encoder to represent the subjective affective information. This enables the learned P-Topics to capture both factual and affective characteristics, as illustrated in Figure \ref{fig:teaser}. We leverage the image's genre and the pair's emotion as proxy labels for the factual and affective aspects to assess the quality of our approach.

We evaluate our approach on two diverse datasets: Firstly, ArtELingo \cite{mohamed2022artelingo} and ArtELingo-28 \cite{mohamed2024no}, which contains $\sim1.2M$ art images with multilingual emotional responses in 28 languages. Secondly, Affection \cite{achlioptas2023affection}, which features realistic images with affective captions in English. This evaluation setting allows us to test our model's ability to capture both fine-grained perceptual differences in artistic content and broader perceptual variations in everyday images.

PercepT produces well-separated (SI 0.97 vs.\ 0.37), semantically grounded clusters (AMI 0.18, V-Measure 0.19 against independent genre and emotion labels), learns effective mapping functions (AUC 0.94 vs.\ 0.77), and is preferred by human evaluators in 58.4\% of pairwise comparisons vs.\ 19.5\% for BERTopic.

Our contributions include:
\begin{itemize}
    \item Formal definition of P-Topic modeling with explicit desiderata for what makes a set of P-Topics practically useful.
    \item \textbf{PercepT}: a two-stage architecture with a DEC-based clustering module and attention-pooling mapping functions.
    \item Extensive evaluation on ArtELingo and Affection, including ablations and human evaluation.
    \item Open-source code, models, and datasets.
\end{itemize}

\section{Related Work}
\label{sec:related}

\noindent
\textbf{Topic Modeling and Text Analysis.}
Topic modeling is fundamental for analyzing document collections \cite{brown-etal-1992-class, blei2003latent}. Recent approaches leverage pretrained language models for nuanced topic discovery \cite{Angelov2020Top2VecDR, sia-etal-2020-tired, meng2022topic}. However, these text-only methods miss the rich information present in visual content.

\noindent\textbf{Multimodal Topic Modeling.}
BERTopic \cite{Grootendorst2022BERTopicNT} enables multimodal modeling using CLIP embeddings \cite{radford2021learning}, but relies on suboptimal off-the-shelf clustering. Subsequent neural approaches like M3L \cite{zosa2022multilingual} and CEMTM \cite{abaskohi2025cemtm} align multimodal embeddings via reconstruction losses or contextualized embeddings from VLLMs \cite{liu2023visual}. In contrast, we propose an Expectation-Maximization-based end-to-end training pipeline for improved performance. Furthermore, while recent prompt-based VLLM methods \cite{reuter2024gptopic, steffen2025more, pham2024topicgpt} extract multimodal topics, they suffer from severe context-length scalability issues on large datasets. Our approach circumvents this by training smaller, task-specific networks.
Beyond architecture, P-Topics modeling introduces a formally distinct problem: existing methods treat topic discovery as an end goal with no mechanism to associate unseen images with discovered topics. P-Topics explicitly separates discovery from mapping, formalizes desiderata for perception experiences, and treats the factual/affective decomposition as a foundational design principle.

\noindent\textbf{Clustering.}
Clustering \cite{jain1988algorithms} is conceptually linked to topic modeling, which can be viewed as soft document clustering \cite{blei2003latent}. Both fields struggle with evaluation without ground-truth labels, requiring a balance between cluster completeness and homogeneity. To address this, we leverage both external (e.g., V-measure \cite{rosenberg2007v}) and internal criteria (e.g., Silhouette Score \cite{rousseeuw1987silhouettes}) for comprehensive validation.

\noindent\textbf{Vision-Language Models.}
Recent works \cite{Yang2022LanguageIA, Kazmierczak2023CLIPQDAAE, Bhalla2024InterpretingCW} have explored language-guided concept bottlenecks and CLIP-based models for explainability, but primarily focus on image classification. This paper instead focuses on modeling diverse perception patterns.

\begin{figure*}[t]
    \centering
    \includegraphics[width=1.0\linewidth]{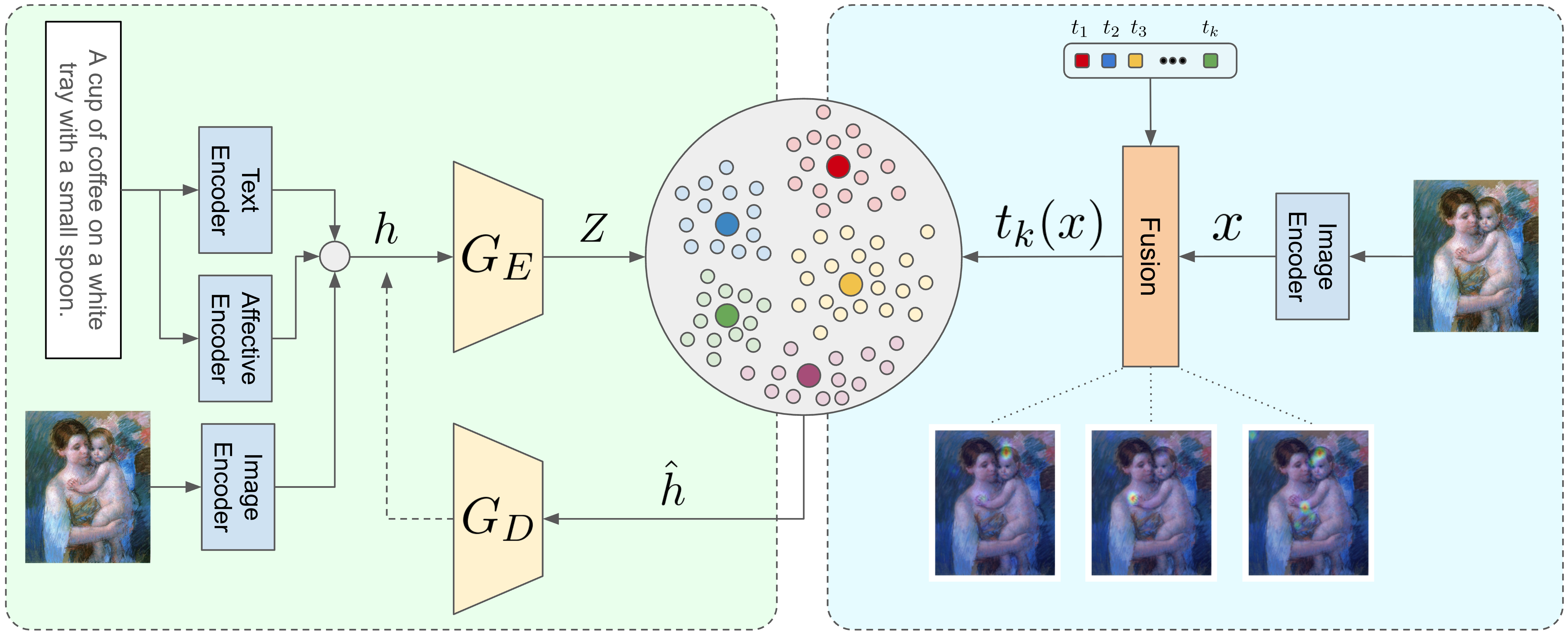}
    \caption{\textbf{PercepT} has two stages: (1) P-Topic Formation (left), where an autoencoder ($G_E, G_D$) learns a latent space $Z$ with distinct clusters (representing P-Topics) from fused embeddings $h$. (2) P-Topic Mapping Functions (right), which maps a new encoded image $x$ to these P-Topics using trainable topic vectors $t_k(x)$.}
    \label{fig:model}
    \vspace{-0.3cm}
\end{figure*}


\noindent\textbf{Cross-Cultural Visual Perception.}
How cultural backgrounds influence visual perception remains under-explored. Despite the availability of datasets like ArtELingo-28 \cite{mohamed2024no, mohamed2022artelingo}, CulturalVQA \cite{nayak-etal-2024-benchmarking}, and GlobalRG \cite{bhatia2024local} few computational models explicitly account for cultural differences in visual interpretation—a notable gap given the increasing need for culturally aware AI systems. 

\section{P-Topics}
\label{sec:formulation}

We conceptualize perception experiences by decomposing them into an \textbf{objective factual} aspect and a \textbf{subjective affective} aspect. Consequently, a \textbf{Perception Topic (P-Topic)} is fundamentally defined as a cluster of image-caption pairs that share similar factual and affective characteristics, capturing a distinct perception experience. For example, given the mother and baby image in Figure \ref{fig:teaser}, one P-Topic might focus on the objective factual context of the image (e.g., religious themes like Mary and Jesus), whereas another P-Topic may characterize the subjective affective response (e.g., feelings of serenity and peace).

More formally, a P-Topic is a distribution ($\mathbb{C}_i$) of image-caption pairs that are embedded in a space capturing both factual and affective characteristics. The image-caption pair embedding is denoted by $h$. It is assigned to $\mathbb{C}_i$ if it is closer to $\mathbb{C}_i$ than any other cluster $\mathbb{C}_j$. 
To be effective in capturing meaningful perceptual experiences, a desired set of P-Topics $\mathbb{C}_i | \forall i = 1 \cdots k$  is expected to satisfy the following desiderata:
\begin{itemize}
    \item \textbf{Distinctiveness:} P-Topics should reflect different distinct experiences, capturing unique combinations of factual and affective aspects.
    \item \textbf{Image Relevance:} Each P-Topic should attend to informative parts of the image.
    \item \textbf{Coverage:} P-Topics altogether capture all of the perceptual experiences in a dataset. 
    \item \textbf{Semantic Coherence:} Each P-Topic does not group contradictory factual or affective viewpoints.    
\end{itemize}


\noindent We also define a \textbf{P-Topic mapping function} ($t_i$) as a function applied to an image ($x$) that yields a topic vector $t_i(x)$ used to compute the assignment probability to $\mathbb{C}_i$.

\noindent Practically, we model a P-Topic $\mathbb{C}_i$ as a cluster $c_i \in \mathcal{C}$ of conceptually and emotionally related image-caption pairs. Given a dataset $\mathcal{D}$, our task is two-fold: (1) discover distinct \textbf{P-Topics} as clusters $c_i \in \mathcal{C} | \forall i = 1 \cdots k$, and (2) learn \textbf{P-Topic mapping functions} $t_i$ that map images to these clusters.

\section{Methodology}
\label{sec:method}

Figure \ref{fig:model} summarizes PercepT, our proposed two-stage method: (1) {\textbf{P-Topic Formation}} and (2) {\textbf{P-Topic Mapping Functions}}, detailed in the subsections below.

{\subsection{P-Topic Formation}}
We discover P-Topics using a denoising autoencoder trained with a neural clustering loss. The autoencoder serves two purposes: reducing the dimensionality of the embedding space ($H$), and shaping the latent space ($Z$) to be more suitable for clustering. Dimensionality reduction is necessary because high-dimensional spaces make clustering unreliable due to the curse of dimensionality \cite{beyer1999nearest}.
Our autoencoder consists of a feed-forward encoder $G_E$ and a feed-forward decoder $G_D$. $G_E$ takes as input an embedding vector $h \in H$ representing an input sample. $h$ is produced by fusing representations from multiple text and image encoders, discussed in \S\ref{sec:representation_fusion}.

\noindent\paragraph{Pretraining.}
We use a reconstruction loss to pretrain $G_E$ and $G_D$ in particular,
\begin{align}
    z &= G_E(h)  \nonumber \\ 
    \hat{z} &= z + \lambda_{noise} \times \mathcal{N}(0, 1)  \nonumber \\
    \hat{h} &= G_D(\hat{z})  \nonumber \\
    L_R &= ||h - \hat{h}||_2  \nonumber
\end{align}

We set $\lambda_{noise} = 0.1$ following \cite{vincent2008dae} and pretrain for 100 epochs or until convergence.

\begin{figure}
    \centering
    \includegraphics[width=0.8\linewidth]{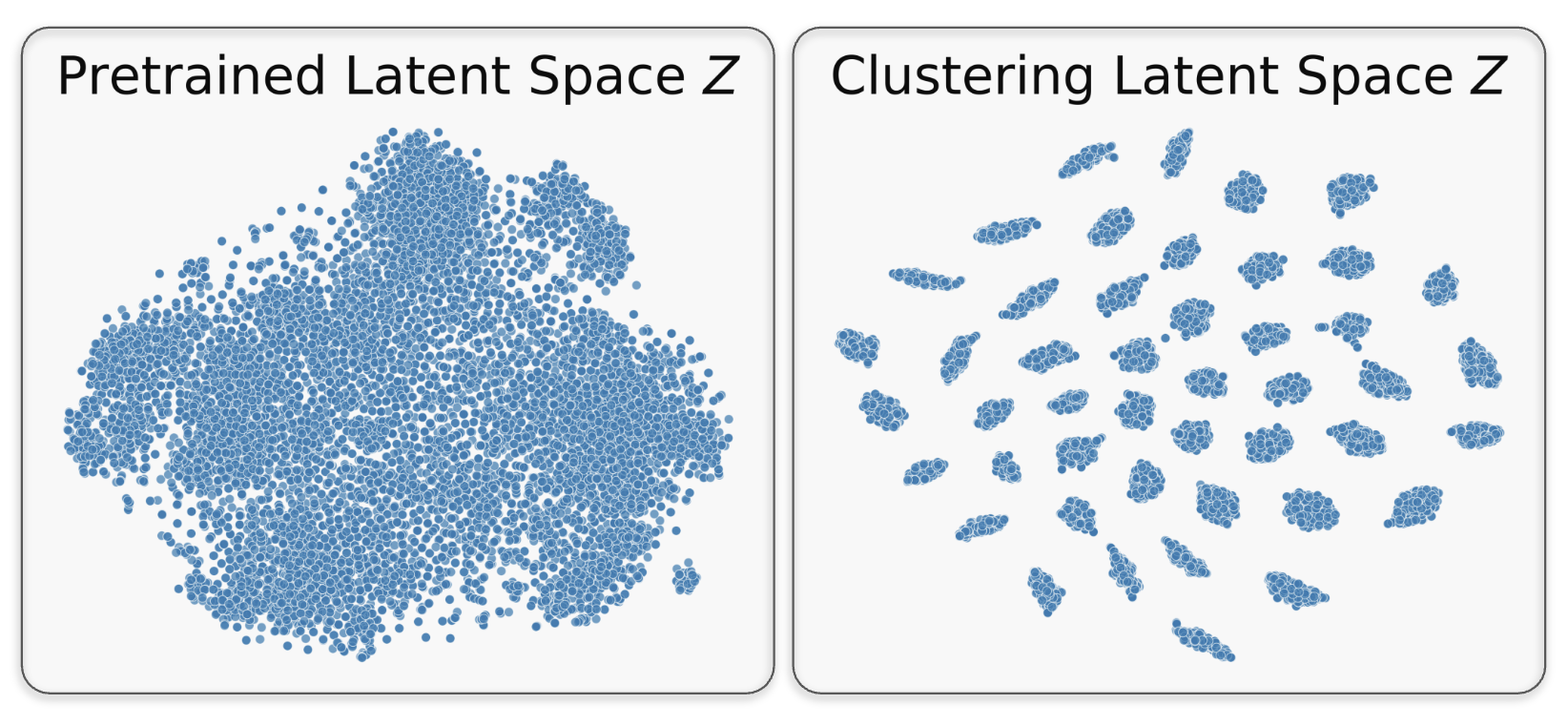}
    \caption{tSNE visualization of the latent space $Z$. Notice how the DEC loss leads to a space better suited for clustering.}
    \label{fig:pretraining_tsne}
    \vspace{-0.5cm}
\end{figure}

After pretraining, $Z$ preserves the structure of $H$ but does not exhibit well-defined clusters, as Figure \ref{fig:pretraining_tsne} (left) shows.

\noindent\paragraph{Neural Clustering Training.}
To improve the clusterability of $Z$, we adopt the DEC loss \cite{xie2016unsupervised} and optimize it via Expectation-Maximization. We initialize $k_0$ trainable cluster centers $\mu_i \in M$, each corresponding to a cluster $c_i \in \mathcal{C}$.

In the Expectation step, we measure soft assignment between latent embeddings and cluster centers (i.e. similarity scores) using the student's $t$-distribution \cite{maaten2008visualizing}:
$$
q_{ij} = \frac{(1 + ||z_i - \mu_j||^2)^{-1}}{\sum_{j'=1}^{k}(1 + ||z_i - \mu_{j'}||^2)^{-1}}
$$
$q_{ij}$ represents the soft assignment probability of embedding $z_i$ to cluster $\mu_j$. To improve distinctiveness and semantic coherence, we minimize the KL-divergence between $Q$ and a sharpened target distribution $P$ \cite{xie2016unsupervised}:
$$
p_{ij} = \frac{q_{ij}^2 / f_j}{\sum_{j'=1}^{k} q_{ij'}^2 / f_{j'}}
$$
where $f_j = \sum_i q_{ij}$ is the soft cluster frequency. 

In the Maximization step, we minimize the KL-divergence between $P$ and $Q$ as follows:
$$
L_{KL} = KL(P||Q) = \sum_i \sum_j p_{ij} \log \frac{p_{ij}}{q_{ij}}
$$

We jointly train $G_E$, $G_D$, and the cluster centers $\mu_i$ by minimizing $L_{KL}$ using backpropagation. To avoid catastrophic forgetting of the reconstruction ability of the autoencoder, we also include the reconstruction loss $L_R$ during this training stage. Thus, our final loss function is 
$
L_{total} = L_{KL} +  \lambda_R L_R
$
We train the model until convergence or for a maximum of 200 epochs.
Figure \ref{fig:pretraining_tsne} shows the tSNE visualization of $Z$ before and after applying the clustering loss. Notice how the clustering loss leads to well-defined clusters in $Z$.

\begin{algorithm}[t]
    \scriptsize
    \caption{\small Finding Threshold Based on Norm's Finite Difference}
    \label{alg:find_threshold}
    \KwIn{Array of norms $\mathbf{norms}$, gap factor $\gamma$}
    \KwOut{Threshold value $\tau$ or \textbf{None}}

    \BlankLine
    Sort $\mathbf{norms}$ in ascending order\;
    Compute $\mathbf{diffs} \leftarrow$ consecutive differences of $\mathbf{norms}$\;
    $g_{\max} \leftarrow \max(\mathbf{diffs})$\;
    \BlankLine
    \If{$g_{\max} > \gamma \cdot \mathrm{mean}(\mathbf{diffs})$}{
        $i^* \leftarrow \arg\max(\mathbf{diffs})$\;
        $\tau \leftarrow \dfrac{\mathbf{norms}[i^*] + \mathbf{norms}[i^* + 1]}{2}$\;
        \Return $\tau$\;
    }
    \Else{
        \Return \textbf{None}\;
    }
\end{algorithm}

\subsection{Original Space Representation Fusion}
\label{sec:representation_fusion}
We fuse representations from multiple pretrained encoders to capture the factual and affective dimensions of P-Topics. For the factual side, we use CLIP-ViT/L-14\footnote{huggingface: openai/clip-vit-large-patch14} \cite{radford2021learning}: its image tower encodes each image $I$ into $h_I \in \mathbb{R}^{768}$ and its text tower encodes each caption $C$ into $h_T \in \mathbb{R}^{768}$. For the affective side, we use ModernBERT-base finetuned on GoEmotions \cite{JdFE2025b} to encode each caption $C$ into an emotional embedding $h_E \in \mathbb{R}^{768}$. We then fuse the three embedding vectors $h_I$, $h_T$, and $h_E$ using two methods: concatenation (projecting $[h_I; h_T; h_E]$ to $d{=}768$ via a learned linear layer) or element-wise weighted mean. The weighted mean fusion scales $h_C$ by $\|h_E\|_2$ and vice versa, steering each representation toward the other's context so that captions evoking different emotions yield distinct $h$ vectors even when their CLIP embeddings are nearly identical. Formally:

\begin{align}
    h_C &= \frac{(h_T + h_I)}{2}  \label{eq:clip_embeds} \\
    h_C' &= h_C * ||h_E||_2 \nonumber \\
    h_E &= h_E * ||h_C||_2 \nonumber \\
    h &= \frac{(2 * h_C' + h_E)}{3}  \label{eq:final_embeds}
\end{align}
We take the mean of $h_I$ and $h_T$ since they share the same CLIP embedding space and have similar magnitude distributions. We weight $h_C'$ with a factor of 2 to reflect the joint image-caption factual signal relative to the single affective signal. We find that this weighted mean fusion leads to better clustering results than concatenation in our experiments.

\begin{figure}
    \centering
    \includegraphics[width=0.8\linewidth]{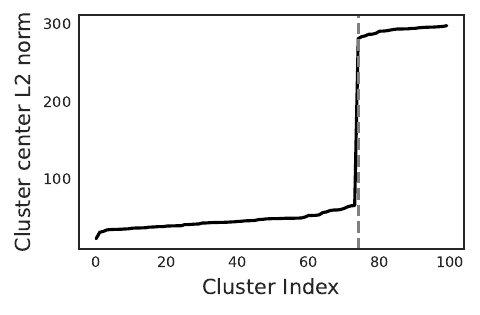}
    \caption{\textbf{Dynamically reducing the number of clusters from 100 to 67.} Clusters with small norms are selected because they reflect perception experiences.}
    \label{fig:dynamic_selection}
\end{figure}

\subsection{Dynamic Selection of Perception Experiences}
We initialize the number of clusters as $k_0$; however, the true number of perception experiences may be fewer. We dynamically filter irrelevant clusters by measuring the norm of each cluster center $\mu_i$: the DEC loss drives centers with few assigned samples to large norms, so we use those norms to discard them. We use Algorithm~\ref{alg:find_threshold} to dynamically set a filtering threshold $\tau$ by finding the point of sharp transition via the maximum finite difference of sorted norms. We then filter out cluster centers with norms larger than $\tau$. Figure \ref{fig:dynamic_selection} shows an example of dynamically selecting the relevant perception experiences using our method. We set $k$ to be the new number of clusters after filtering.

{\subsection{Learning P-Topic Mapping Functions}}
\label{sec:p_topics_modeling}
In the second stage, we aim to learn the P-Topic mapping functions whose task is to map an input image $I$ to a set of identified P-Topic clusters in $\mathcal{C}$. 
This task can be viewed as a multi-label classification problem: the input is the image embeddings $x \in \mathbb{R}^{N\times d}$ ($N$ patches, $d$ embedding dimension), and the output is a binary vector $O \in \mathbb{R}^{k}$ indicating the presence of each perception experience. 

We set $O[i]=1$ if any caption of image $I$ is assigned to cluster $c_i$.

We instantiate each P-Topic mapping function as a trainable topic vector $t_i \in T$ where $|T| = k$ is the number of P-topics. We first pool the image patch embeddings into a shared context vector using attention pooling, then apply a per-topic feed-forward head $F_i$ to produce a topic-specific representation $t_i(x)$, where $x$ is the encoded image from the image encoder. We then calculate an assignment probability score $s_i$ for each topic mapping function $t_i$ as $s_i = \sigma(F_i(t_i(x)))$, where $F_i$ is a feed-forward layer with a single value output. We train the topic vectors $t_i$ and the feed-forward layers $F_i$ to minimize the binary cross-entropy loss $L_{P}$ between the output scores $s_i$ and the ground-truth one-hot vector $O$. During inference, we assign an image $I$ to the perception experience ($c_i$) if $s_i$ exceeds a threshold $\delta$.
More formally $L_{P}$ is calculated as: 
\begin{align}
    t_i(x) &= \text{AttnPool}(x) \\
    s_i &= \sigma(F_i(t_i(x))) \\
    L_P &= -\frac{1}{k} \sum_{i=1}^{k} \Big(
            O[i]\log(s_i) + \nonumber \\
        &    \quad\quad (1 - O[i])\log(1 - s_i)
        \Big)
\end{align}

\noindent where $\sigma$ is the sigmoid function and the exact implementation of $\text{AttnPool}$ can be found in the supplementary materials.
We name this approach \textbf{PercepT} in our experiments. Alternatively, we ablate using a simpler linear layer model, named \textbf{PercepT$_{Linear}$} and a Multi-Headed attention layer, named \textbf{PercepT$_{Multi}$}. Our ablations show that \textbf{PercepT} performs the best. 

\section{Experiments}
\label{sec:experiments}

\subsection{Datasets}
\noindent\textbf{ArtELingo.} We combine ArtELingo \cite{mohamed2022artelingo} and ArtELingo-28 \cite{mohamed2024no} to form a training set of $\sim 1.2M$ image-caption pairs in 28 languages. Each sample includes a WikiArt
\footnote{wikiart organization} 
image, an emotional caption, an emotion label, and an image genre. The genre and emotion reflect factual and affective perspectives, respectively, letting us evaluate if PercepT captures both. To prevent our text encoder from latching onto script-specific surface features, we translate all non-English captions to English using GemmaX2\footnote{huggingface: ModelSpace/GemmaX2-28-9B-v0.1} \cite{cui2025multilingualmachinetranslationopen}. This does not discard culturally relevant information: the subjective perspectives and emotional responses originate from annotators spanning 28 languages and are preserved in translation.

\noindent\textbf{Affection.} We also use the Affection dataset \cite{achlioptas2023affection}, containing $\sim 500K$ realistic image-caption pairs reflecting human emotions but lacking genre labels. Since results on Affection mirror ArtELingo, we defer them to the supplementary materials due to space constraints.

\begin{figure*}[t]
    \centering
    \includegraphics[width=0.9\linewidth]{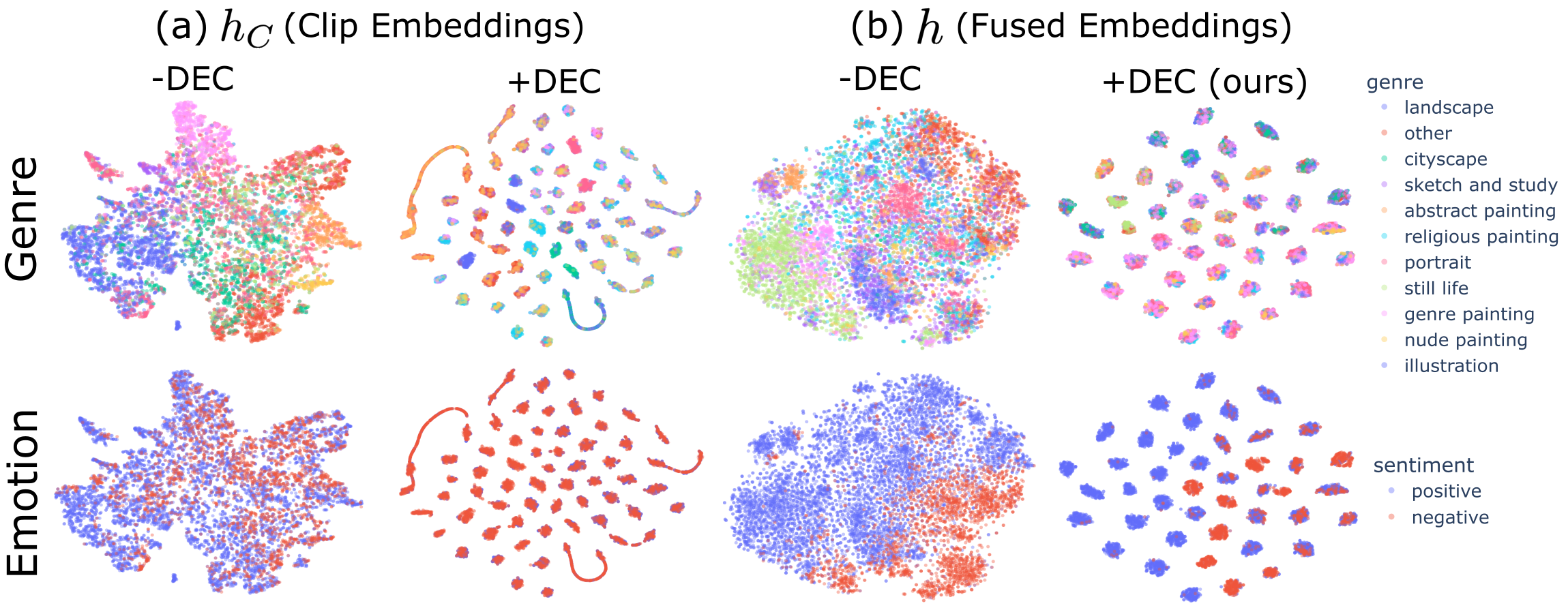} 
    \caption{\textbf{t-SNE projection of the learned latent space on ArtELingo.} The DEC loss significantly results in more distinct, well-organized clusters. Meanwhile the fused embeddings $h$ show better alignment with both genre and emotion. PercepT is the rightmost model.}
    \label{fig:tsne_labels}
    \vspace{-0.4cm}
\end{figure*}

\subsection{Baselines}
We compare PercepT against several baselines that utilize different representation learning and clustering methods.
\begin{itemize}
    \item \textbf{BERTopic} \cite{Grootendorst2022BERTopicNT}: We adapt BERTopic to our problem by utilizing its multimodal backend to build multimodal topics.
    \item \textbf{CEMTM} \cite{abaskohi2025cemtm}: We train a CEMTM model on our datasets using the same procedure in the original paper.
    \item \textbf{VLM2VEC} \cite{jiang2024vlm2vec}: We utilize both pretrained versions (v1 and v2) to evaluate VLLMs representations on our task. 
    \item \textbf{K-Means on CLIP Embeddings $h_C$}: We extract $h_C$ (eq. \ref{eq:clip_embeds}) for each image-caption pair, followed by dimensionality reduction, then K-Means clustering to form the clusters. We named them $CLIP$.
    \item \textbf{K-Means on Fused Embeddings $h$}: same as the above but we use the fused embeddings $h$ (eq. \ref{eq:final_embeds}) instead. We named them $CLIP+Emo$.
\end{itemize}
For dimensionality reduction, we compare both PCA \cite{tipping1999probabilistic} and SVD \cite{halko2011finding}. We did not observe any significant differences between them so we report only PCA in table \ref{tab:clustering_metrics}.

\subsection{Evaluation Metrics}
For Stage 1 we use silhouette score (\textbf{SI}) \cite{rousseeuw1987silhouettes}, Davies-Bouldin index (\textbf{DBI}) \cite{davies2009cluster}, and Calinski-Harabasz index (\textbf{CHI}) \cite{calinski1974dendrite} to measure cluster structure. To evaluate alignment with genre and emotion labels, we use Rand score (\textbf{ARI}) \cite{rand1971objective}, adjusted mutual information (\textbf{AMI}) \cite{JMLR:v11:vinh10a}, homogeneity (\textbf{H}), completeness (\textbf{C}), and V-Measure (\textbf{V}) \cite{rosenberg2007v}. We also employ human evaluators to assess cluster quality and ensure its alignment with human judgement. 
For the second stage, we report standard multi-label classification metrics: AUC, precision, recall, F1-score, and accuracy. We treat \textbf{AUC as the primary metric}, as it measures ranking quality across all possible thresholds. Precision, recall, F1, and accuracy are threshold-dependent and therefore sensitive to the choice of $\delta$, making them secondary indicators.
Since the ground-truth labels for Stage 2 are derived from the Stage 1 cluster assignments, these metrics reflect the distinctiveness and coherence of Stage 1 clusters. The more separated the clusters from Stage 1 and the more coherent they are, the easier the classification task in Stage 2 becomes.
Qualitatively, we inspect the clusters visually using t-SNE \cite{maaten2008visualizing}.
These metrics serve as proxies for the desiderata defined in \S\ref{sec:formulation}: SI, CHI, and DBI measure \textbf{Distinctiveness}; AMI, V-Measure, and human evaluation measure \textbf{Semantic Coherence}; multi-label AUC and F1 measure \textbf{Coverage}.

\subsection{P-Topics Formation Results}
\subsubsection{Qualitative Evaluation -- Latent Space Visualization.}
\textit{How well does an unsupervised clustering method such as PercepT agree with labels such as genre and emotion?} Figure \ref{fig:tsne_labels} shows the t-SNE projection of the learned latent space, with colors encoding genre (top row) and emotion (bottom row). The left and right columns compare $h_C$ (CLIP embeddings) and $h$ (fused embeddings). DEC results in more distinct clusters in both cases. While $h_C$ captures the factual aspect (genre) well, it fails on the affective aspect (emotion); $h$ captures both.

\begin{table*}[t]
    \centering
    \small
    \renewcommand{\arraystretch}{1}
    \begin{tabular}{l|l|c|ccc|cccccc}
        \toprule
        & \multicolumn{2}{l|}{\textbf{Method}} & \textbf{SI} $\uparrow$ & \textbf{CHI} $\uparrow$ & \textbf{DBI} $\downarrow$ & \textbf{ARI} $\uparrow$ & \textbf{AMI} $\uparrow$ & \textbf{H} $\uparrow$ & \textbf{C} $\uparrow$ & \textbf{V} $\uparrow$ & \textbf{FMI} $\uparrow$ \\
        \hline
        \rowcolor{gray!20} \cellcolor{white} & \multicolumn{2}{l|}{BERTopic-MM} & 0.37 & 125757 & 0.98 & 0.01 & 0.07 & 0.12 & 0.05 & 0.07 & 0.07 \\
        \rowcolor{gray!20} \cellcolor{white} & \multicolumn{2}{l|}{CEMTM} & 0.20 & 3227 & 0.81 & 0.01 & 0.01 & 0.04 & 0.03 & 0.03 & 0.10 \\
        \rowcolor{gray!20} \cellcolor{white} & \multicolumn{2}{l|}{VLM2VEC v1} & 0.03 & 44 & 3.05 & 0.02 & 0.16 & 0.26 & 0.14 & 0.18 & 0.09 \\
        \rowcolor{gray!20} \cellcolor{white} & \multicolumn{2}{l|}{VLM2VEC v2} & 0.03 & 40 & 3.19 & 0.01 & 0.11 & 0.23 & 0.10 & 0.14 & 0.08 \\
        \rowcolor{gray!20} \cellcolor{white} & \multicolumn{2}{l|}{$CLIP$} & 0.02 & 1309 & 3.23 & 0.01 & 0.09 & 0.15 & 0.06 & 0.09 & 0.08 \\
        \rowcolor{gray!20} \cellcolor{white} \multirow{-6}{*}{{\rotatebox{90}{\textbf{baselines}}}} & \multicolumn{2}{l|}{$CLIP+Emo$} & 0.02 & 1236.09 & 3.35 & 0.02 & 0.14 & 0.24 & 0.10 & 0.14 & 0.09 \\
        \hline
        \rowcolor{yellow!20} \cellcolor{white}\textbf{ours} & \multicolumn{2}{l|}{\textbf{PercepT ($\lambda_R=1$)}} & \textbf{0.97} & \textbf{291574} & \underline{0.33} & \textbf{0.04} & \textbf{0.18} & \underline{0.30} & \textbf{0.14} & \textbf{0.19} & \textbf{0.12} \\
        \hline
        \multirow{9}{*}{\rotatebox{90}{\textbf{ablations}}}
        & \multicolumn{2}{l|}{\quad $\lambda_R = 0.1$} & 0.83 & 16447 & 0.43 & 0.03 & 0.15 & 0.20 & 0.12 & 0.15 & 0.10 \\
        & \multicolumn{2}{l|}{\quad $\lambda_R = 10$} & 0.71 & 35006 & 0.37 & 0.03 & \textbf{0.18} & 0.29 & \textbf{0.14} & \textbf{0.19} & 0.10 \\
        & \multicolumn{2}{l|}{\quad $\lambda_R = 100$} & 0.58 & 32058 & 0.55 & 0.02 & 0.16 & 0.25 & 0.12 & 0.16 & 0.09 \\
        \cline{2-12}
        & \textbf{Embedding} & \textbf{DEC Loss} & & & & & & & & & \\
        \cline{2-12}
        & $h_C$ & \checkmark & \underline{0.91} & \underline{146645} & \textbf{0.15} & 0.01 & 0.05 & 0.07 & 0.03 & 0.05 & 0.08 \\
        & $h_{concat}$ & \checkmark & 0.83 & 8153 & 0.53 & 0.01 & 0.03 & 0.04 & 0.02 & 0.03 & 0.10 \\
        \cline{2-12}
        & $h_C$ & - & 0.13 & 179 & 1.51 & 0.01 & 0.08 & 0.12 & 0.06 & 0.08 & 0.09 \\
        & $h$ & - & 0.02 & 13 & 3.09 & \underline{0.03} & 0.16 & \textbf{0.32} & \textbf{0.14} & \textbf{0.19} & \underline{0.10} \\
        & $h_{concat}$ & - & 0.09 & 39 & 2.41 & 0.01 & 0.07 & 0.11 & 0.05 & 0.07 & 0.07 \\
        \hline
        \bottomrule
    \end{tabular}
    \caption{\textbf{Quantitative clustering evaluation.} PercepT (\textbf{yellow}) 
    clearly outperforms BERTopic, the second best baseline. PercepT demonstrates superior performance in both internal cluster structure (e.g., \textbf{Silhouette 0.97}) and external semantic alignment (\textbf{V-Measure 0.19}). The ablation shows that DEC loss is critical for forming dense, well-separated clusters (see $h$ without DEC). PercepT corresponds to $h$ with DEC.}
    \label{tab:clustering_metrics}
    \vspace{-0.3cm}
\end{table*}

\subsubsection{Quantitative Evaluation}
\noindent\textbf{Clustering Metrics.} Table \ref{tab:clustering_metrics} presents the quantitative results for the clustering stage. PercepT, corresponding to \textbf{$\boldsymbol{h}$ $+$DEC}, significantly outperforms all baselines across both internal metrics (cluster structure) and external metrics (label alignment). Notably, PercepT achieves the highest Silhouette score (0.97) and Calinski-Harabasz index (291574). These scores are substantially better than BERTopic (0.37) and orders of magnitude higher than the standard K-Means approaches (e.g., $CLIP+Emo$ Silhouette 0.02). This indicates that our model produces exceptionally dense and well-separated clusters, a finding that quantitatively corroborates the t-SNE visualization in Figure \ref{fig:tsne_labels}. A high SI is the \emph{expected} outcome of DEC; the critical question is whether the resulting clusters are also semantically meaningful.
For the external metrics, we report the harmonic mean between the emotion and genre scores. PercepT achieves top scores in AMI (0.18), V-Measure (0.19), and ARI (0.04) against independent genre and emotion labels that were never seen during training, confirming that the well-separated clusters capture genuine perceptual content instead of being degenerate solutions. 

\noindent\textbf{Ablation.} The ablation study reveals three critical insights. (1) The DEC loss is essential for cluster structure. The model using our fused embeddings without DEC, h ($-$DEC), suffers a catastrophic drop in internal metrics (Silhouette 0.02), producing a poorly structured latent space. (2) Our fused embeddings are essential to capture the dual nature of perception experiences. Applying DEC directly to raw CLIP embeddings, $h_C$ ($+$DEC), creates structurally sound clusters (Silhouette 0.91) that have little correlation with the ground-truth labels (AMI 0.05, V-Measure 0.05). This demonstrates that while $h_C$ lacks the necessary subjective information to align with emotions, our fusion module successfully injects it into $h$. (3) The optimal $\lambda_R = 1$ where we find the scores decrease for both $\lambda_R = 0.1$ and $\lambda_R = 10$
        
\begin{figure}[t]
    \vspace{-0.3cm}
    \centering
    \includegraphics[width=0.7\linewidth]{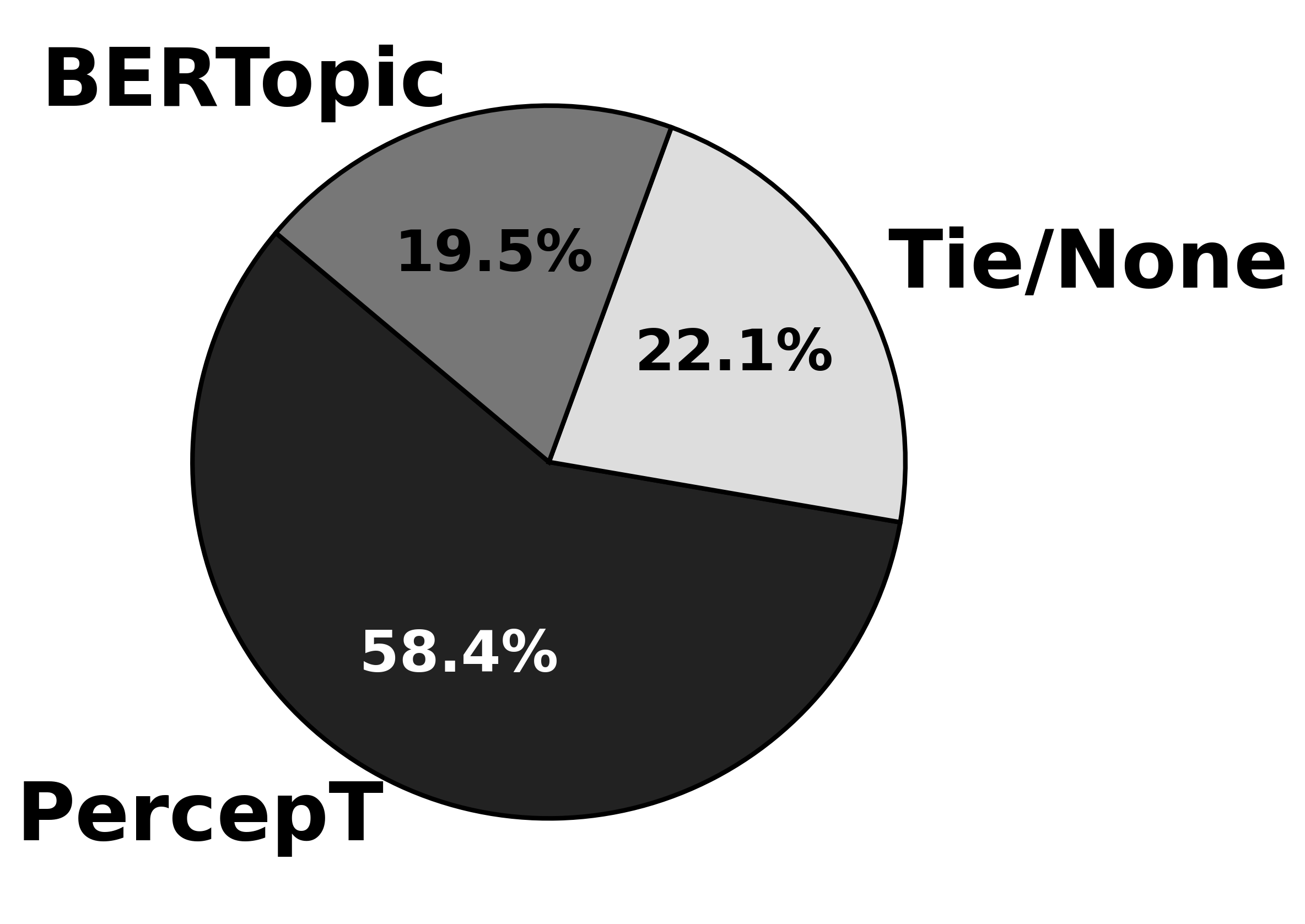}
    \caption{Human evaluation results for PercepT and BERTopic.}
    \label{fig:human_evaluation}
    \vspace{-0.5cm}
\end{figure}

\noindent\textbf{Human Evaluation.} To complement our quantitative metrics, we conducted a human evaluation comparing PercepT and BERTopic clusters. As shown in Figure \ref{fig:human_evaluation}, evaluators preferred PercepT in $58.4\%$ of cases vs.\ $19.5\%$ for BERTopic, with $22.1\%$ ties or neither. We include the collection interface and annotator statistics in the appendix.

\begin{table}[t]
    \centering
    \small
    \begin{tabular}{l|ccccc}
        \toprule
        \textbf{Model} & \textbf{AUC} & \textbf{F1} & \textbf{P} & \textbf{R} & \textbf{Acc} \\
        \hline
    
        \rowcolor{yellow!20} \textbf{PercepT} & \textbf{0.94} & \textbf{0.61} & \textbf{0.67} & \textbf{0.56} & \textbf{0.06} \\
        PercepT$_{Linear}$ & 0.93 & 0.59 & 0.65 & 0.54 & 0.04 \\
        PercepT$_{Multi}$ & 0.91 & 0.57 & 0.59 & 0.55 & 0.03 \\
        \hline
        \hline
    
        \rowcolor{gray!20} BERTopic & 0.77 & 0.38 & 0.35 & 0.41 & 0.00 \\
        \rowcolor{gray!20} PCA      & 0.86 & 0.45 & 0.42 & 0.49 & 0.01 \\
        \hline
    
        \bottomrule
    \end{tabular}
    \caption{\textbf{Quantitative P-Topics Evaluation on ArtELingo.} PercepT achieves highest scores across multi-label classification metrics, significantly outperforming the \textbf{BERTopic} baseline.}
    \label{tab:ptopics_metrics}
    \vspace{-0.5cm}
\end{table}

\subsection{P-Topics Mapping Results}

\noindent\textbf{Quantitative Evaluation.} Table \ref{tab:ptopics_metrics} details P-Topic mapping performance on ArtELingo. The well-separated, coherent clusters produced by Stage 1 provide clear and unambiguous targets, making the mapping task tractable. PercepT with Attention Pooling achieves the best results (AUC: 0.94, F1: 0.61), effectively capturing underlying perception themes by focusing on relevant image tokens. It substantially outperforms the BERTopic baseline (AUC: 0.77, F1: 0.38). PCA utilizes the learned latent space from Stage 1, which explains why it outperforms BERTopic. Nonetheless, PercepT significantly outperforms PCA as well, highlighting the importance of our DEC loss in forming clusters that are easier to learn from.

\noindent The low exact-match accuracy scores are expected: with $k=67$ labels, a single wrong label scores zero and predictions rarely achieve an exact match. AUC is therefore the primary metric, measuring assignment quality across all thresholds.

\noindent\textbf{Ablation.} The Linear variant performs surprisingly well, suggesting the high-quality clusters from Stage 1 simplify P-Topic learning. Conversely, the minor drop with Multi-Headed attention indicates that excessive complexity is unnecessary.  

\section{Conclusion}
\label{sec:conclusion}

We introduced \textbf{P-Topics modeling} and \textbf{PercepT}, a two-stage architecture that discovers perception experiences as DEC-based clusters over fused factual and affective embeddings, then learns attention-pooling mapping functions to attribute unseen images to those experiences.
Evaluations on ArtELingo and Affection set a new state-of-the-art: Silhouette \textbf{0.97} and AUC \textbf{0.94}, with human evaluators overwhelmingly preferring PercepT's clusters. By jointly modeling the factual and affective dimensions of visual perception, our method paves the way for more culturally aware and emotionally intelligent vision-language systems.

\section*{Limitations}

\noindent\textbf{Image Relevance evaluation.} Our desiderata include Image Relevance — whether each P-Topic attends to informative image regions — but we do not measure it quantitatively. Developing spatial attention maps or region-level grounding metrics for P-Topics remains an open direction.

\noindent\textbf{Translation of multilingual captions.} To prevent script-specific artifacts in the text encoder, we translate all non-English captions to English. While cultural content is largely preserved, machine translation may smooth over culturally specific phrasings and emotion expressions, potentially reducing the cultural signal available during clustering.

\noindent\textbf{Caption quality dependency.} Stage 1 clustering relies on captions that accurately reflect the annotator's perceptual experience. In datasets with minimal, generic, or noisy captions, the affective signal in $h_E$ would be weaker, and the quality of the discovered P-Topics would degrade accordingly.

\noindent\textbf{Dataset scope.} We evaluate on art images (ArtELingo) and realistic images with affective captions (Affection). Generalization to other visual domains — e.g., news photography, social media, or medical imagery — remains to be established.

\bibliography{custom}

@String(CVPR  = {IEEE Conf. Comput. Vis. Pattern Recog.})

@String(AAAI  = {AAAI})

@String(CVPR  = {CVPR})

@String(CVPR= {IEEE Conf. Comput. Vis. Pattern Recog.})

@String(AAAI = {AAAI})

@article{Bhalla2024InterpretingCW,
  author    = {Usha Bhalla and Alexander X. Oesterling and Suraj Srinivas and F. Calmon and Himabindu Lakkaraju},
  booktitle = {Neural Information Processing Systems},
  journal   = {ArXiv},
  title     = {Interpreting {CLIP} with Sparse Linear Concept Embeddings (SpLiCE)},
  volume    = {abs/2402.10376},
  year      = {2024}
}

@article{Kazmierczak2023CLIPQDAAE,
  title   = {{CLIP-QDA}: An Explainable Concept Bottleneck Model},
  author  = {R{\'e}mi Kazmierczak and Eloise Berthier and Goran Frehse and Gianni Franchi},
  journal = {Trans. Mach. Learn. Res.},
  year    = {2023},
  volume  = {2024},
  url     = {https://api.semanticscholar.org/CorpusID:265552032}
}

@article{brown-etal-1992-class,
  title   = {Class-Based \textit{n}-gram Models of Natural Language},
  author  = {Brown, Peter F.  and
             Della Pietra, Vincent J.  and
             deSouza, Peter V.  and
             Lai, Jenifer C.  and
             Mercer, Robert L.},
  journal = {Computational Linguistics},
  volume  = {18},
  number  = {4},
  year    = {1992},
  url     = {https://aclanthology.org/J92-4003/},
  pages   = {467--480}
}

@article{Angelov2020Top2VecDR,
  title   = {{Top2Vec}: Distributed Representations of Topics},
  author  = {Dimitar Angelov},
  journal = {ArXiv},
  year    = {2020},
  volume  = {abs/2008.09470},
  url     = {https://api.semanticscholar.org/CorpusID:221246303}
}

@inproceedings{sia-etal-2020-tired,
  title     = {Tired of Topic Models? Clusters of Pretrained Word Embeddings Make for Fast and Good Topics too!},
  author    = {Sia, Suzanna  and
               Dalmia, Ayush  and
               Mielke, Sabrina J.},
  editor    = {Webber, Bonnie  and
               Cohn, Trevor  and
               He, Yulan  and
               Liu, Yang},
  booktitle = {Proceedings of the 2020 Conference on Empirical Methods in Natural Language Processing (EMNLP)},
  month     = nov,
  year      = {2020},
  address   = {Online},
  publisher = {Association for Computational Linguistics},
  url       = {https://aclanthology.org/2020.emnlp-main.135/},
  doi       = {10.18653/v1/2020.emnlp-main.135},
  pages     = {1728--1736}
}

@article{Grootendorst2022BERTopicNT,
  title   = {{BERTopic}: Neural topic modeling with a class-based TF-IDF procedure},
  author  = {Maarten R. Grootendorst},
  journal = {ArXiv},
  year    = {2022},
  volume  = {abs/2203.05794},
  url     = {https://api.semanticscholar.org/CorpusID:247411231}
}

@inproceedings{meng2022topic,
  title     = {Topic discovery via latent space clustering of pretrained language model representations},
  author    = {Meng, Yu and Zhang, Yunyi and Huang, Jiaxin and Zhang, Yu and Han, Jiawei},
  booktitle = {Proceedings of the ACM Web Conference 2022},
  pages     = {3143--3152},
  year      = {2022}
}

@article{Yang2022LanguageIA,
  author    = {Yue Yang and Artemis Panagopoulou and Shenghao Zhou and Daniel Jin and Chris Callison-Burch and Mark Yatskar},
  booktitle = {Computer Vision and Pattern Recognition},
  journal   = {2023 IEEE/CVF Conference on Computer Vision and Pattern Recognition (CVPR)},
  pages     = {19187-19197},
  title     = {Language in a Bottle: Language Model Guided Concept Bottlenecks for Interpretable Image Classification},
  year      = {2022}
}

@article{mohamed2024no,
  title     = {No Culture Left Behind: {A}rt{EL}ingo-28, a Benchmark of {W}iki{A}rt with Captions in 28 Languages},
  author    = {Mohamed, Youssef  and
               Li, Runjia  and
               Ahmad, Ibrahim Said  and
               Haydarov, Kilichbek  and
               Torr, Philip  and
               Church, Kenneth  and
               Elhoseiny, Mohamed},
  editor    = {Al-Onaizan, Yaser  and
               Bansal, Mohit  and
               Chen, Yun-Nung},
  booktitle = {Proceedings of the 2024 Conference on Empirical Methods in Natural Language Processing},
  journal   = {Proceedings of the 2024 Conference on Empirical Methods in Natural Language Processing},
  month     = nov,
  year      = {2024},
  address   = {Miami, Florida, USA},
  publisher = {Association for Computational Linguistics},
  url       = {https://aclanthology.org/2024.emnlp-main.1165/},
  doi       = {10.18653/v1/2024.emnlp-main.1165},
  pages     = {20939--20962}
}

@article{blei2003latent,
  title   = {Latent dirichlet allocation},
  author  = {Blei, David M and Ng, Andrew Y and Jordan, Michael I},
  journal = {Journal of machine Learning research},
  volume  = {3},
  number  = {Jan},
  pages   = {993--1022},
  year    = {2003}
}

@article{mohamed2022artelingo,
  title   = {Artelingo: A million emotion annotations of wikiart with emphasis on diversity over language and culture},
  author  = {Mohamed, Youssef and Abdelfattah, Mohamed and Alhuwaider, Shyma and Li, Feifan and Zhang, Xiangliang and Church, Kenneth Ward and Elhoseiny, Mohamed},
  journal = {arXiv preprint arXiv:2211.10780},
  year    = {2022}
}

@article{elhoseiny2016write,
  title={Write a classifier: Predicting visual classifiers from unstructured text},
  author={Elhoseiny, Mohamed and Elgammal, Ahmed and Saleh, Babak},
  journal={IEEE transactions on pattern analysis and machine intelligence},
  volume={39},
  number={12},
  pages={2539--2553},
  year={2016},
  publisher={IEEE}
}

@inproceedings{elhoseiny2013write,
  title={Write a classifier: Zero-shot learning using purely textual descriptions},
  author={Elhoseiny, Mohamed and Saleh, Babak and Elgammal, Ahmed},
  booktitle={Proceedings of the IEEE international conference on computer vision},
  pages={2584--2591},
  year={2013}
}

@inproceedings{mohamed2022okay,
  title     = {It is okay to not be okay: Overcoming emotional bias in affective image captioning by contrastive data collection},
  author    = {Mohamed, Youssef and Khan, Faizan Farooq and Haydarov, Kilichbek and Elhoseiny, Mohamed},
  booktitle = {Proceedings of the IEEE/CVF Conference on Computer Vision and Pattern Recognition},
  pages     = {21263--21272},
  year      = {2022}
}

@inproceedings{beyer1999nearest,
  title        = {When is “nearest neighbor” meaningful?},
  author       = {Beyer, Kevin and Goldstein, Jonathan and Ramakrishnan, Raghu and Shaft, Uri},
  booktitle    = {International conference on database theory},
  pages        = {217--235},
  year         = {1999},
  organization = {Springer}
}

@inproceedings{vincent2008dae,
  title        = {Extracting and composing robust features with denoising autoencoders},
  author       = {Vincent, Pascal and Larochelle, Hugo and Bengio, Yoshua and Manzagol, Pierre-Antoine},
  booktitle    = {Proceedings of the 25th international conference on Machine learning},
  pages        = {1096--1103},
  year         = {2008},
  organization = {ACM}
}

@inproceedings{xie2016unsupervised,
  title        = {Unsupervised deep embedding for clustering analysis},
  author       = {Xie, Junyuan and Girshick, Ross and Farhadi, Ali},
  booktitle    = {International conference on machine learning},
  pages        = {478--487},
  year         = {2016},
  organization = {PMLR}
}

@article{maaten2008visualizing,
  title   = {Visualizing data using t-SNE},
  author  = {Maaten, Laurens van der and Hinton, Geoffrey},
  journal = {Journal of machine learning research},
  volume  = {9},
  number  = {Nov},
  pages   = {2579--2605},
  year    = {2008}
}

@inproceedings{radford2021learning,
  title        = {Learning transferable visual models from natural language supervision},
  author       = {Radford, Alec and Kim, Jong Wook and Hallacy, Chris and Ramesh, Aditya and Goh, Gabriel and Agarwal, Sandhini and Sastry, Girish and Askell, Amanda and Mishkin, Pamela and Clark, Jack and others},
  booktitle    = {International conference on machine learning},
  pages        = {8748--8763},
  year         = {2021},
  organization = {PmLR}
}

@inproceedings{JdFE2025b,
  title        = {Emotion Detection with ModernBERT},
  author       = {Enric Junqu\'e de Fortuny},
  year         = {2025},
  howpublished = {\url{https://huggingface.co/cirimus/modernbert-base-go-emotions}}
}

@misc{cui2025multilingualmachinetranslationopen,
  title         = {Multilingual Machine Translation with Open Large Language Models at Practical Scale: An Empirical Study},
  author        = {Menglong Cui and Pengzhi Gao and Wei Liu and Jian Luan and Bin Wang},
  year          = {2025},
  eprint        = {2502.02481},
  archiveprefix = {arXiv},
  primaryclass  = {cs.CL},
  url           = {https://arxiv.org/abs/2502.02481}
}

@inproceedings{achlioptas2023affection,
  title     = {Affection: Learning affective explanations for real-world visual data},
  author    = {Achlioptas, Panos and Ovsjanikov, Maks and Guibas, Leonidas and Tulyakov, Sergey},
  booktitle = {Proceedings of the IEEE/CVF conference on computer vision and pattern recognition},
  pages     = {6641--6651},
  year      = {2023}
}

@article{tipping1999probabilistic,
  title     = {Probabilistic principal component analysis},
  author    = {Tipping, Michael E and Bishop, Christopher M},
  journal   = {Journal of the Royal Statistical Society Series B: Statistical Methodology},
  volume    = {61},
  number    = {3},
  pages     = {611--622},
  year      = {1999},
  publisher = {Oxford University Press}
}

@article{rousseeuw1987silhouettes,
  title     = {Silhouettes: a graphical aid to the interpretation and validation of cluster analysis},
  author    = {Rousseeuw, Peter J},
  journal   = {Journal of computational and applied mathematics},
  volume    = {20},
  pages     = {53--65},
  year      = {1987},
  publisher = {Elsevier}
}

@article{davies2009cluster,
  title     = {A cluster separation measure},
  author    = {Davies, David L and Bouldin, Donald W},
  journal   = {IEEE transactions on pattern analysis and machine intelligence},
  number    = {2},
  pages     = {224--227},
  year      = {2009},
  publisher = {Ieee}
}

@article{calinski1974dendrite,
  title     = {A dendrite method for cluster analysis},
  author    = {Cali{\'n}ski, Tadeusz and Harabasz, Jerzy},
  journal   = {Communications in Statistics-theory and Methods},
  volume    = {3},
  number    = {1},
  pages     = {1--27},
  year      = {1974},
  publisher = {Taylor \& Francis}
}

@article{rand1971objective,
  title     = {Objective criteria for the evaluation of clustering methods},
  author    = {Rand, William M},
  journal   = {Journal of the American Statistical association},
  volume    = {66},
  number    = {336},
  pages     = {846--850},
  year      = {1971},
  publisher = {Taylor \& Francis}
}

@article{JMLR:v11:vinh10a,
  author  = {Nguyen Xuan Vinh and Julien Epps and James Bailey},
  title   = {Information Theoretic Measures for Clusterings Comparison: Variants, Properties, Normalization and Correction for Chance},
  journal = {Journal of Machine Learning Research},
  year    = {2010},
  volume  = {11},
  number  = {95},
  pages   = {2837--2854},
  url     = {http://jmlr.org/papers/v11/vinh10a.html}
}

@inproceedings{rosenberg2007v,
  title     = {V-measure: A conditional entropy-based external cluster evaluation measure},
  author    = {Rosenberg, Andrew and Hirschberg, Julia},
  booktitle = {Proceedings of the 2007 joint conference on empirical methods in natural language processing and computational natural language learning (EMNLP-CoNLL)},
  pages     = {410--420},
  year      = {2007}
}

@book{jain1988algorithms,
  title     = {Algorithms for clustering data},
  author    = {Jain, Anil K and Dubes, Richard C},
  year      = {1988},
  publisher = {Prentice-Hall, Inc.}
}

@article{halko2011finding,
  title     = {Finding structure with randomness: Probabilistic algorithms for constructing approximate matrix decompositions},
  author    = {Halko, Nathan and Martinsson, Per-Gunnar and Tropp, Joel A},
  journal   = {SIAM review},
  volume    = {53},
  number    = {2},
  pages     = {217--288},
  year      = {2011},
  publisher = {SIAM}
}

@misc{qwen3technicalreport,
  title         = {Qwen3 Technical Report},
  author        = {Qwen Team},
  year          = {2025},
  eprint        = {2505.09388},
  archiveprefix = {arXiv},
  primaryclass  = {cs.CL},
  url           = {https://arxiv.org/abs/2505.09388}
}

@inproceedings{abaskohi2025cemtm,
  title     = {CEMTM: Contextual Embedding-based Multimodal Topic Modeling},
  author    = {Abaskohi, Amirhossein and Li, Raymond and Li, Chuyuan and Joty, Shafiq and Carenini, Giuseppe},
  booktitle = {Proceedings of the 2025 Conference on Empirical Methods in Natural Language Processing},
  pages     = {11686--11703},
  year      = {2025}
}

@inproceedings{zosa2022multilingual,
  title     = {Multilingual and multimodal topic modelling with pretrained embeddings},
  author    = {Zosa, Elaine and Pivovarova, Lidia},
  booktitle = {Proceedings of the 29th International Conference on Computational Linguistics},
  pages     = {4037--4048},
  year      = {2022}
}

@article{reuter2024gptopic,
  title   = {Gptopic: Dynamic and interactive topic representations},
  author  = {Reuter, Arik and Khadka, Bishnu and Thielmann, Anton and Weisser, Christoph and Fischer, Sebastian and S{\"a}fken, Benjamin},
  journal = {arXiv preprint arXiv:2403.03628},
  year    = {2024}
}

@inproceedings{pham2024topicgpt,
  title     = {TopicGPT: A prompt-based topic modeling framework},
  author    = {Pham, Chau Minh and Hoyle, Alexander and Sun, Simeng and Resnik, Philip and Iyyer, Mohit},
  booktitle = {Proceedings of the 2024 Conference of the North American Chapter of the Association for Computational Linguistics: Human Language Technologies (Volume 1: Long Papers)},
  pages     = {2956--2984},
  year      = {2024}
}

@inproceedings{steffen2025more,
  title     = {More than memes: A multimodal topic modeling approach to conspiracy theories on telegram},
  author    = {Steffen, Elisabeth},
  booktitle = {Proceedings of the International AAAI Conference on Web and Social Media},
  volume    = {19},
  pages     = {1831--1844},
  year      = {2025}
}

@article{liu2023visual,
  title   = {Visual instruction tuning},
  author  = {Liu, Haotian and Li, Chunyuan and Wu, Qingyang and Lee, Yong Jae},
  journal = {Advances in neural information processing systems},
  volume  = {36},
  pages   = {34892--34916},
  year    = {2023}
}

@article{jiang2024vlm2vec,
  title   = {Vlm2vec: Training vision-language models for massive multimodal embedding tasks},
  author  = {Jiang, Ziyan and Meng, Rui and Yang, Xinyi and Yavuz, Semih and Zhou, Yingbo and Chen, Wenhu},
  journal = {arXiv preprint arXiv:2410.05160},
  year    = {2024}
}

@inproceedings{nayak-etal-2024-benchmarking,
    title = "Benchmarking Vision Language Models for Cultural Understanding",
    author = "Nayak, Shravan  and
      Jain, Kanishk  and
      Awal, Rabiul  and
      Reddy, Siva  and
      Steenkiste, Sjoerd Van  and
      Hendricks, Lisa Anne  and
      Stanczak, Karolina  and
      Agrawal, Aishwarya",
    editor = "Al-Onaizan, Yaser  and
      Bansal, Mohit  and
      Chen, Yun-Nung",
    booktitle = "Proceedings of the 2024 Conference on Empirical Methods in Natural Language Processing",
    month = nov,
    year = "2024",
    address = "Miami, Florida, USA",
    publisher = "Association for Computational Linguistics",
    url = "https://aclanthology.org/2024.emnlp-main.329/",
    doi = "10.18653/v1/2024.emnlp-main.329",
    pages = "5769--5790"
}

@article{bhatia2024local,
  title={From Local Concepts to Universals: Evaluating the Multicultural Understanding of Vision-Language Models},
  author={Bhatia, Mehar and Ravi, Sahithya and Chinchure, Aditya and Hwang, Eunjeong and Shwartz, Vered},
  journal={arXiv preprint arXiv:2407.00263},
  year={2024}
}

\clearpage

\appendix

\begin{table*}[t]
    \centering
    \small
    \setlength{\tabcolsep}{3pt}
    \renewcommand{\arraystretch}{1.2}
\begin{tabular}{l|l|c|ccc|cccccc}
    \toprule
    & \multicolumn{2}{l|}{\textbf{Method}} & \textbf{SI} $\uparrow$ & \textbf{CHI} $\uparrow$ & \textbf{DBI} $\downarrow$ & \textbf{ARI} $\uparrow$ & \textbf{AMI} $\uparrow$ & \textbf{H} $\uparrow$ & \textbf{C} $\uparrow$ & \textbf{V} $\uparrow$ & \textbf{FMI} $\uparrow$ \\
    \hline
    \rowcolor{gray!20} \cellcolor{white} & \multicolumn{2}{l|}{BERTopic-MM} & 0.54 & 328195.59 & 0.64 & 0.01 & 0.07 & 0.11 & 0.05 & 0.07 & 0.07 \\
    \rowcolor{gray!20} \cellcolor{white} & \multicolumn{2}{l|}{$CLIP$} & 0.06 & 2702.99 & 2.94 & 0.01 & 0.08 & 0.13 & 0.06 & 0.08 & 0.06 \\
    \rowcolor{gray!20} \cellcolor{white} \multirow{-3}{*}{\rotatebox{90}{\tiny\textbf{baselines}}} & \multicolumn{2}{l|}{$CLIP + EMO$} & 0.05 & 2511.79 & 3.09 & 0.01 & 0.11 & 0.17 & 0.08 & 0.11 & 0.07 \\
    \hline
    \rowcolor{yellow!20} \cellcolor{white}\textbf{ours} & \multicolumn{2}{l|}{\textbf{PercepT} ($\lambda_R = 1$)} & 0.93 & 261452.72 & 0.1 & 0.03 & 0.18 & 0.28 & 0.13 & 0.18 & 0.1 \\
    \hline
    & \multicolumn{2}{l|}{\quad $\lambda_R = 0.1$} & 0.41 & 10946.51 & 0.99 & 0.0 & 0.04 & 0.04 & 0.05 & 0.04 & 0.19 \\
    & \multicolumn{2}{l|}{\quad $\lambda_R = 10$} & 0.65 & 105303.6 & 0.42 & 0.01 & 0.05 & 0.08 & 0.04 & 0.05 & 0.06 \\
    & \multicolumn{2}{l|}{\quad $\lambda_R = 100$} & 0.47 & 97231.99 & 0.59 & 0.01 & 0.03 & 0.05 & 0.02 & 0.03 & 0.06 \\
    \cline{2-12}
    & \textbf{Embedding} & \textbf{DEC Loss} & & & & & & & & & \\
    \cline{2-12}
    & $h_C$ & \checkmark & 0.96 & 392106.76 & 0.31 & 0.01 & 0.06 & 0.09 & 0.04 & 0.06 & 0.06 \\
    & $h_{concat}$ & \checkmark & 0.76 & 72132.21 & 0.33 & 0.03 & 0.18 & 0.3 & 0.13 & 0.18 & 0.1 \\
    \cline{2-12}
    & $h_C$ & - & 0.15 & 283.71 & 1.39 & 0.01 & 0.07 & 0.11 & 0.06 & 0.07 & 0.08 \\
    & $h$ & - & 0.03 & 34.86 & 3.58 & 0.04 & 0.21 & 0.33 & 0.16 & 0.21 & 0.13 \\
    \multirow{-9}{*}{\rotatebox{90}{\textbf{ablations}}} & $h_{concat}$ & - & 0.17 & 295.66 & 1.36 & 0.05 & 0.25 & 0.39 & 0.18 & 0.25 & 0.13 \\
    \hline
    \bottomrule
\end{tabular}
    \caption{\textbf{Quantitative clustering evaluation on Affection dataset.}}
    \label{tab:clustering_metrics_affection}
\end{table*}

\begin{table}[t!]
    \scriptsize
    \centering
    \begin{tabular}{l|ccccc}
        \toprule
        \textbf{Model} & \textbf{AUC} & \textbf{F1} & \textbf{P} & \textbf{R} & \textbf{Acc} \\
        \hline
    
        \rowcolor{yellow!20} \textbf{PercepT} & 0.89 & 0.28 & 0.58 & 0.19 & 0.02 \\
        PercepT$_{Linear}$ & 0.88 & 0.28 & 0.54 & 0.18 & 0.02 \\
        PercepT$_{Multi-Head}$ & 0.84 & 0.31 & 0.32 & 0.31 & 0.01 \\
        \hline
        \hline
    
        \rowcolor{gray!20} BERTopic & 0.87 & 0.41 & 0.57 & 0.35 & 0.06 \\
        \rowcolor{gray!20} PCA      & 0.88 & 0.41 & 0.56 & 0.37 & 0.11  \\
        \hline
    
        \bottomrule
    \end{tabular}
    \caption{\textbf{Quantitative P-Topics Evaluation on Affection.}}
    \label{tab:ptopics_metrics_affection}
\end{table}

\section{P-Topics Architecture Details}
We compared the performance of three different architectures: (1) Multi-Headed Attention, (2) Attention Pooling, and (3) Linear. PercepT uses Attention Pooling since it outperforms the other variants. We provide the details for each architecture below.

\subsection{Multi-Headed Attention}
The core of our architecture is the mapping of an input image to a set of learnable P-Topics. Let $\mathbf{X} \in \mathbb{R}^{B \times N \times D_1}$ denote the encoded image embeddings (where $N$ is the number of patches) and $\mathbf{T} \in \mathbb{R}^{B \times K \times D_2}$ denote the learnable topic vectors, where $K$ is the number of perception clusters. $B$, $D_1$ and $D_2$ are the batch sizes, image embedding dimensionality, and P-Topics dimensionality, respectively. 

The P-Topic vectors $\mathbf{T}$ act as queries, while the image embeddings $\mathbf{X}$ serve as the context (keys and values).

First, the inputs are projected into query, key, and value embeddings:
\begin{align}
    \mathbf{Q} &= \mathbf{T}\mathbf{W}_Q \\
    \mathbf{K} &= \mathbf{X}\mathbf{W}_K, \quad \mathbf{V} = \mathbf{X}\mathbf{W}_V
\end{align}
where $\mathbf{W}_Q, \mathbf{W}_K, \mathbf{W}_V$ transform the inputs to the inner dimension $D_{inner} = H \cdot d_k$.
Here, $\mathbf{Q}$ represents the topic queries and $\mathbf{K}, \mathbf{V}$ represent the image context.

The raw similarity scores $\mathbf{S} \in \mathbb{R}^{B \times H \times K \times N}$ are computed via scaled dot-product:
\begin{equation}
    \mathbf{S}_{h,i,j} = \frac{1}{\sqrt{d_k}} (\mathbf{Q}_{h,i} \cdot \mathbf{K}_{h,j}^\top)
\end{equation}

\paragraph{Talking Heads Mixing.}
To capture complex interactions between different perceptual perspectives, we employ a ``talking heads'' mechanism. This allows information flow between attention heads before the normalization step. Let $\mathbf{\Theta} \in \mathbb{R}^{H \times H}$ be a learnable mixing matrix:
\begin{equation}
    \mathbf{S}'_h = \sum_{k=1}^H \mathbf{\Theta}_{h,k} \mathbf{S}_k
\end{equation}

Finally, The attention weights are then computed via softmax along the image patch dimension, and the attended topic representations $\mathbf{T}_{out}$ are aggregated:
\begin{align}
    \mathbf{A} &= \text{Dropout}(\text{Softmax}(\mathbf{S}', \text{dim}=-1)) \\
    \mathbf{O}_h &= \mathbf{A}_h \mathbf{V}_h \\
    \mathbf{T}_{out} &= \text{Concat}(\mathbf{O}_1, \dots, \mathbf{O}_H) \mathbf{W}_O
\end{align}
This results in topic-specific image representations $\mathbf{T}_{out} \in \mathbb{R}^{B \times K \times D}$, denoted as $t_i(x)$ in the P-Topic formalism. $D$ is the latent space dimensionality.

\subsection{Attention Pooling}
To aggregate the feature sequence into a fixed-size representation, we utilize an Attention Pooling module. Given the encoded image embeddings $\mathbf{X} \in \mathbb{R}^{B \times N \times D_1}$, we first compute unnormalized attention logits $\mathbf{E} \in \mathbb{R}^{B \times N \times 1}$ via a learnable linear projection $\mathbf{W}_{att} \in \mathbb{R}^{D_1 \times 1}$:
\begin{equation}
    \mathbf{E} = \mathbf{X}\mathbf{W}_{att}
\end{equation}

These logits are normalized across the sequence dimension $N$ using the Softmax function to produce attention weights $\bm{\alpha} \in \mathbb{R}^{B \times N \times 1}$:
\begin{equation}
    \bm{\alpha} = \text{Softmax}(\mathbf{E}, \text{dim}=1)
\end{equation}

We then compute the weighted sum of the input features to produce a pooled context vector $\mathbf{v} \in \mathbb{R}^{B \times D_{in}}$:
\begin{equation}
    \mathbf{v} = \sum_{n=1}^{N} \bm{\alpha}_{n} \mathbf{X}_{n}
\end{equation}

The pooled vector is projected to the latent space output dimension $D_{out}$ using weights $\mathbf{W}_{proj} \in \mathbb{R}^{D_1 \times K}$, where $K$ is the number of perception clusters. Finally, we apply a Sigmoid activation function to calculate the assignment probability:
\begin{equation}
    \mathbf{S} = \sigma(\mathbf{v}\mathbf{W}_{proj})
\end{equation}
Where $\mathbf{S} \in \mathbb{R}^{B \times K}$ represents the assignment probability to the perception clusters.

\subsection{Linear}
As a baseline, we utilize a direct linear transformation. Given an image feature vector $\mathbf{x} \in \mathbb{R}^{B \times D_1}$, the output $\mathbf{y} \in \mathbb{R}^{B \times K}$ is computed via an affine transformation:
\begin{equation}
    \mathbf{y} = \mathbf{x}\mathbf{W}_{linear} + \mathbf{b} \\
    \mathbf{S} = \sigma(\mathbf{y})
\end{equation}
where $\mathbf{W}_{linear} \in \mathbb{R}^{D_1 \times D_K}$ represents the learnable weight matrix and $\mathbf{b} \in \mathbb{R}^{D_K}$ denotes the bias term. Meanwhile $\mathbf{S} \in \mathbb{R}^{B \times K}$ represents the assignment probability to the perception clusters.

Notice that here we do not use patch features from the input image. Instead, we use a single vector corresponding to the \texttt{[CLS]} token of the image encoder to represent the image.

\section{Affection Dataset Results}
Table \ref{tab:clustering_metrics_affection} shows the clustering metrics results on the Affection dataset. We observe the same trends observed in table \ref{tab:clustering_metrics} in the main paper.
Similarly, the P-Topics stage evaluation results are found in table \ref{tab:ptopics_metrics_affection}.

\paragraph{Analysis of Stage 2 Results on Affection.}
Table~\ref{tab:ptopics_metrics_affection} shows that PercepT obtains a lower F1 score (0.28) than BERTopic (0.41) on Affection, which warrants explanation.

\noindent\textbf{F1 is threshold-sensitive; AUC is not.}
F1 is computed at a fixed assignment threshold of $\delta=0.5$. This default may not be optimal for every model and dataset combination, so F1 comparisons at a single threshold can be misleading. AUC integrates performance across all possible thresholds and is therefore a more reliable indicator of intrinsic ranking quality. PercepT achieves a higher AUC (0.89) than BERTopic (0.87), confirming that at the optimal operating threshold PercepT would outperform BERTopic in F1 as well.

\noindent\textbf{Affection lacks the perceptual diversity that benefits PercepT.}
Affection was collected entirely from English-speaking annotators, producing captions that reflect a narrower range of cultural and affective perspectives than ArtELingo, which spans 28 languages and cultures. PercepT's core advantage lies in disentangling the factual and affective dimensions of perception: when the underlying dataset already contains limited affective diversity, the benefit of our fused embeddings and DEC-based clustering is reduced, and simpler baselines become comparably competitive. The larger gains on ArtELingo (PercepT AUC 0.94 vs.\ BERTopic 0.77) versus the narrower margin on Affection (0.89 vs.\ 0.87) are consistent with this interpretation.

\section{Qualitative Analysis of P-Topics}

In our final model, each P-Topic corresponds to a perception experience cluster. Each cluster can be represented by the datapoints assigned to it. We represent each cluster using a combination of image–caption pairs and the most salient words. For the image–caption pairs, we use the assignment score as a probability and sample $N$ pairs from the assigned image–caption pairs. For the salient words, we utilize Qwen3-14B\footnote{\url{huggingface: Qwen/Qwen3-14B}} \cite{qwen3technicalreport} to generate the most important words given all the captions in a cluster. We use the prompt below:

\begin{quote}
\begin{verbatim}
Analyze the following sentences and 
extract exactly 10 keywords that best
represent the main themes, objects, 
and concepts across all sentences.

Sentences:
{combined_text}

Instructions:
- Provide exactly 10 single words 
  or short phrases (1–2 words)
- Mention the common objects in 
  the sentences
- Choose words that capture the 
  core themes
- Return only the words, 
  separated by commas
- Do not number them or add 
  explanations
- Avoid vague or overly general
  terms

Keywords:
\end{verbatim}
\end{quote}

where \texttt{\{combined\_text\}} is replaced with a list of all captions belonging to the cluster.

For the visualization in Figure \ref{fig:teaser}, we utilize the wordcloud package\footnote{pypi: project/wordcloud} to generate the word cloud for the keywords of each cluster.

\section{BERTopic Multimodal Adaptation Details}
\label{sec:bertopic_details}

To ensure a fair and reproducible comparison, we document the exact configuration used to adapt BERTopic~\cite{Grootendorst2022BERTopicNT} to our multimodal P-Topics setting.

\paragraph{Multimodal Embeddings.}
We use BERTopic's built-in multimodal backend, which accepts precomputed image embeddings in place of text embeddings. Images are encoded with \textbf{CLIP ViT-B/32} (\texttt{openai/clip-vit-base-patch32})~\cite{radford2021learning}. This differs from the CLIP ViT-L/14 encoder used in PercepT; we chose ViT-B/32 because BERTopic's multimodal pipeline was designed and validated with this model, so substituting a larger backbone would introduce confounds beyond the clustering algorithm itself. Captions are encoded with BERTopic's default sentence-transformer backbone.

\paragraph{Dimensionality Reduction.}
BERTopic's default UMAP~\cite{mcinnes2018umap} reducer is applied to the joint image–text embeddings prior to clustering. We retain BERTopic's default UMAP hyperparameters (\texttt{n\_neighbors=15}, \texttt{n\_components=5}, \texttt{metric=`cosine'}).

\paragraph{Cluster Count.}
We set \texttt{nr\_topics=100} to match the initial number of cluster centers used in PercepT's P-Topic Formation stage, ensuring both methods start with the same target granularity. BERTopic's HDBSCAN step first finds its own number of topics, which are then merged down to 100 via BERTopic's topic reduction procedure.

\paragraph{Summary.}
Table~\ref{tab:bertopic_config} summarises the configuration for easy reference.

\begin{table}[h]
    \centering
    \scriptsize
    \begin{tabular}{ll}
        \toprule
        \textbf{Component} & \textbf{Setting} \\
        \midrule
        Image encoder      & CLIP ViT-B/32 \\
        Text encoder       & BERTopic default (MiniLM) \\
        Dimensionality reduction & UMAP (default params) \\
        Clustering         & HDBSCAN (default params) \\
        Target cluster count (\texttt{nr\_topics}) & 100 \\
        \bottomrule
    \end{tabular}
    \caption{BERTopic multimodal adaptation configuration.}
    \label{tab:bertopic_config}
\end{table}

\section{Implementation Details}

\paragraph{Model Architecture.}
We implement our framework using PyTorch. The backbone consists of a Clustering Autoencoder with a symmetric encoder-decoder architecture. The encoder is composed of Multi-Layer Perceptrons (MLP) with Rectified Linear Unit (ReLU) activations. Based on our configuration, the encoder hidden layers have dimensions $[500, 500, 500, 2000]$, projecting the input embeddings into a latent feature space of dimension $d=128$. The decoder mirrors this structure to reconstruct the original input. We set the number of latent clusters (topics) to $K=100$. The cluster centers are initialized using K-Means on the latent features of the pretrained autoencoder. For the mapping from image embeddings to topics (P-Topics), we utilize a projection layer followed by a Sigmoid activation to output probability scores.

\paragraph{Training Protocol.}
Our training pipeline consists of two distinct stages: autoencoder pretraining and topic mapping training.

\begin{itemize}
    \item \textbf{Autoencoder Pretraining:} We first pretrain the autoencoder to minimize the Mean Squared Error (MSE) reconstruction loss. We train for $100$ epochs using the Adam optimizer with a learning rate of $1 \times 10^{-3}$. We employ a Cosine Annealing learning rate scheduler, decaying the learning rate to a minimum of $1 \times 10^{-5}$ over the course of training. The batch size is set to $512$.
    
    \item \textbf{Deep Clustering (Joint Training):} Following pretraining, the model undergoes joint training to refine both the feature representations and cluster assignments. We optimize a combined loss function consisting of the reconstruction loss and the clustering loss (Kullback-Leibler divergence between soft assignments and the target distribution). The cluster assignments are computed using a Student's $t$-distribution kernel. We use the same Adam optimizer with a learning rate of $1 \times 10^{-3}$ as well as the same Cosine Annealing scheduler. We train this stage for $200$ epochs using a batch size of $512$
    
    \item \textbf{Topic Mapping (P-Topics):} Finally, we freeze the text-derived clusters and train the visual mapping module. This stage is trained for $100$ epochs with a batch size of $64$ images. We use the Binary Cross Entropy (BCE) loss to align the predicted visual topic probabilities with the ground-truth cluster labels derived from the text captions. The optimizer and scheduler settings remain consistent with the pretraining phase (Adam, lr=$1 \times 10^{-3}$, Cosine Annealing).
\end{itemize}

\section{Human Evaluation}
\begin{figure}
    \centering
    \includegraphics[width=1\linewidth]{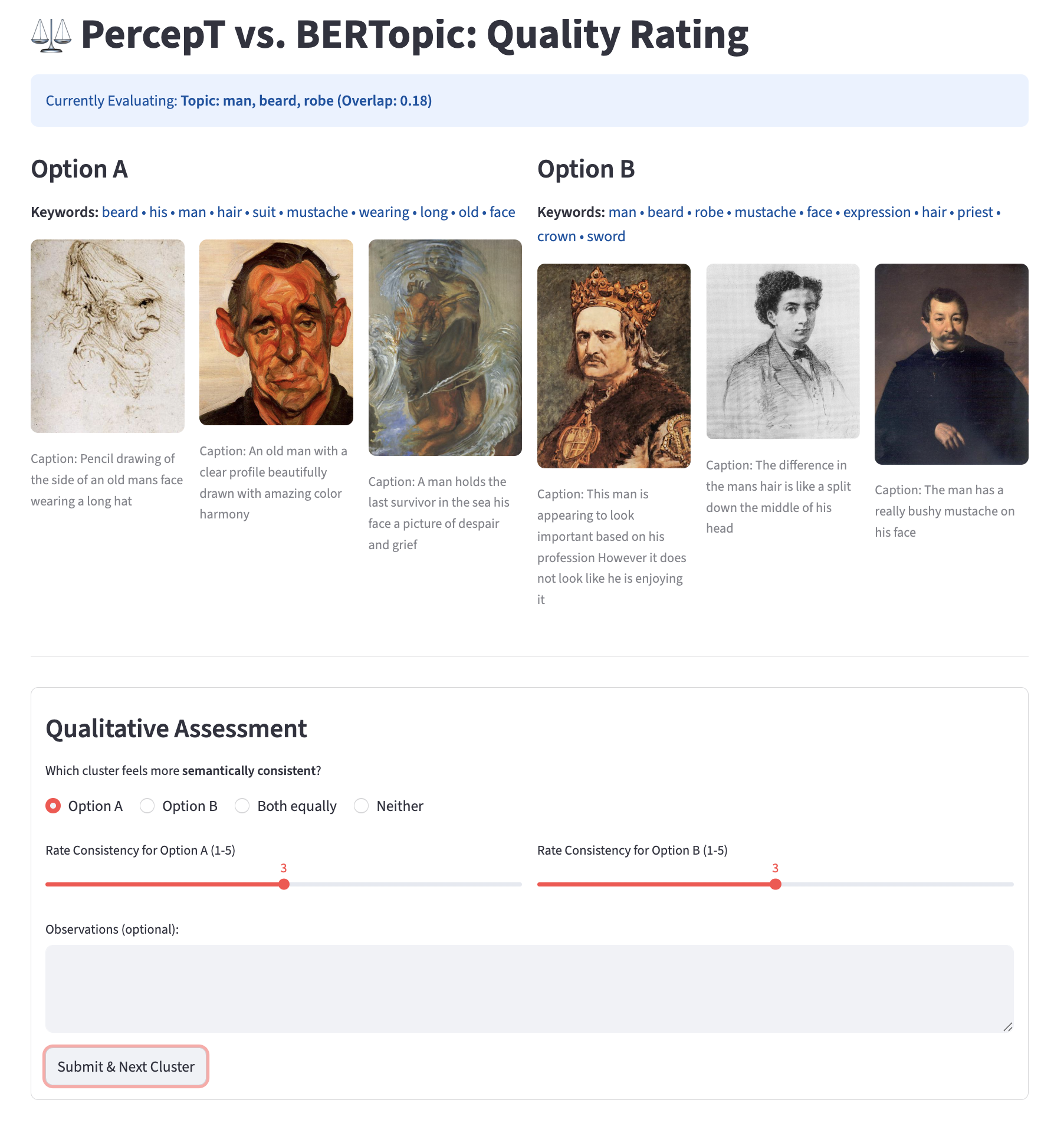}
    \caption{\textbf{Human Evaluation Interface.} The annotator is asked to select which model has better consistency and represents the P-Topic keywords better.}
    \label{fig:human_setup}
\end{figure}

We assess the quality of the P-Topics formed by PercepT and BERTopic (the second best model) via a model vs. model comparison. The interface we used for the collection can be found in figure \ref{fig:human_setup}. We match the clusters from BERTopic and PercepT using the Jaccard similarity between the keywords. We then anonymize the model name randomly to be either model A or model B. Finally, the models are placed randomly either on the left or right panes. The annotators are then asked to select the best model. We also add options for both good or neither is good. The evaluation was performed by 4 CS PhD students aging between 20-30 years. They consist of 3 males and 1 female. We collected a total of 113 samples. In addition, we collected consistency rates for each model.

\end{document}